\documentclass[11pt,letterpaper]{article}
\usepackage{naaclhlt2016}
\usepackage{times}
\usepackage{latexsym}
\usepackage{amssymb}

\usepackage{amsmath}
\usepackage{amsthm}
\usepackage{amssymb}
\usepackage{graphicx}
\usepackage{xspace}
\usepackage{tabularx}
\usepackage{multicol}
\usepackage{multirow}
\usepackage{url}
\usepackage{wrapfig}
\usepackage{comment}
\usepackage{listings}
\usepackage{color}
\usepackage[utf8]{inputenc}
\usepackage{fancyvrb}
\usepackage{booktabs}
\usepackage{color}
\usepackage[normalem]{ulem}

\newcommand{\bmx}[0]{\begin{bmatrix}}
\newcommand{\emx}[0]{\end{bmatrix}}

\newcommand{\vect}[1]{\mathbf{#1}}

\newcommand{\matr}[1]{\mathbf{#1}}

\newcommand{\vc}[0]{\vect{c}}
\newcommand{\ve}[0]{\vect{e}}

\newcommand{\vh}[0]{\vect{h}}

\newcommand{\vx}[0]{\vect{x}}

\newcommand{\vw}[0]{\vect{w}}

\newcommand{\vf}[0]{\vect{f}}
\newcommand{\vi}[0]{\vect{i}}
\newcommand{\vo}[0]{\vect{o}}

\newcommand{\vg}[0]{\vect{g}}

\newcommand{\vr}[0]{\vect{r}}

\newcommand{\mW}[0]{\matr{W}}

\newcommand{\mU}[0]{\matr{U}}
\newcommand{\mF}[0]{\matr{F}}

\newcommand{\RR}[0]{\mathbb{R}}

\newcommand{\ola}{\overleftarrow}
\newcommand{\ora}{\overrightarrow}

\naaclfinalcopy 
\def\naaclpaperid{***} 

\title{Efficient Character-level Document Classification by  \\ Combining Convolution and Recurrent Layers}

\author{Yijun Xiao\\
		Center for Data Sciences, \\
	    New York University\\
	    {\tt ryjxiao@nyu.edu}
	  \And
	Kyunghyun Cho\\
    Courant Institute and \\
    Center for Data Science, \\
  	New York University\\
  {\tt kyunghyun.cho@nyu.edu}}

\date{}

\begin{document}

\maketitle

\begin{abstract}
Document classification tasks were primarily tackled at word level. Recent research that works with character-level inputs shows several benefits over word-level approaches such as natural incorporation of morphemes and better handling of rare words. We propose a neural network architecture that utilizes both convolution and recurrent layers to efficiently encode character inputs. We validate the proposed model on eight large scale document classification tasks and compare with character-level convolution-only models. It achieves comparable performances with much less parameters.
\end{abstract}

\section{Introduction}

Document classification is a task in natural language processing where one needs
to assign a single or multiple predefined categories to a sequence of text.
A conventional approach to document classification generally consists of a
feature extraction stage followed by a classification stage. For instance, it is
usual to use a TF-IDF vector of a given document as an input feature to a
subsequent classifier.

More recently, it has become more common to use a deep neural network, which
jointly performs feature extraction and classification, for document
classification
\cite{kim2014convolutional,mesnil2014ensemble,socher2013recursive,Carrier2014}.
In most cases, an input document is represented as a sequence of words, of which
each is presented as a one-hot vector.\footnote{
    A one-hot vector of the $i$-th word is a binary vector whose elements are
    all zeros, except for the $i$-th element which is set to one.
} Each word in the sequence is projected into a continuous vector space by being
multiplied with a weight matrix, forming a sequence of dense, real-valued
vectors. This sequence is then fed into a deep neural network which processes
the sequence in multiple layers, resulting in a prediction probability.
This whole pipeline, or a network, is tuned jointly to maximize the
classification accuracy on a training set.

One important aspect of these recent approaches based on deep learning is that
they often work at the level of words. Despite its recent success, the
word-level approach has a number of major shortcomings. First, it is
statistically inefficient, as each word token is considered separately and
estimated by the same number of parameters, despite the fact that many words
share common root, prefix or suffix. This can be overcome by using an external
mechanism to segment each word and infer its components (root, prefix, suffix),
but this is not desirable as the mechanism is highly language-dependent and is
tuned independently from the target objective of document classification. 

Second, the word-level approach cannot handle out-of-vocabulary words. Any word
that is not present or rare in a training corpus, is mapped to an unknown word
token. This is problematic, because the model cannot handle typos easily, which
happens frequently in informal documents such as postings from social network
sites. Also, this makes it difficult to use a trained model to a new domain, as
there may be large mismatch between the domain of the training corpus and the
target domain.

Recently this year, a number of researchers have noticed that it is not at all
necessary for a deep neural network to work at the word level. As long as the
document is represented as a sequence of one-hot vectors, the model works
without any change, regardless of whether each one-hot vector corresponds to a
word, a sub-word unit or a character. Based on this intuition, Kim et
al.~\cite{kim2015character} and Ling et al.~\cite{ling2015finding} proposed to
use a character sequence as an alternative to the word-level one-hot vector. A
similar idea was applied to dependency parsing in
\cite{ballesteros2015improved}. The work in this direction, most relevant to
this paper, is the character-level convolutional network for document
classification by Zhang et al.~\cite{zhang2015character}.

The character-level convolutional net in \cite{zhang2015character} is composed
of many layers of convolution and max-pooling, similarly to the convolutional
network in computer vision (see, e.g., \cite{NIPS2012_4824}.) Each layer first
extracts features from small, overlapping windows of the input sequence and
pools over small, non-overlapping windows by taking the maximum activations in
the window. This is applied recursively (with untied weights) for many times.
The final convolutional layer's activation is flattened to form a vector which
is then fed into a small number of fully-connected layers followed by the
classification layer.

We notice that the use of a vanilla convolutional network for character-level
document classification has one shortcoming. As the receptive field of each
convolutional layer is often small (7 or 3 in \cite{zhang2015character},) the
network must have many layers in order to capture long-term dependencies in an
input sentence. This is likely the reason why Zhang et
al.~\cite{zhang2015character} used a very deep convolutional network with six
convolutional layers followed by two fully-connected layers.

In order to overcome this inefficiency in modeling a character-level sequence,
in this paper we propose to make a hybrid of convolutional and recurrent
networks. This was motivated by recent successes of applying recurrent
networks to natural languages (see, e.g.,
\cite{cho2014learning,sundermeyer2015feedforward}) and from the fact that the
recurrent network can efficiently capture long-term dependencies even with a
single layer. The hybrid model processes an input sequence of
characters with a number of convolutional layers followed by a single recurrent
layer. Because the recurrent layer, consisting of either gated recurrent units
(GRU, \cite{cho2014learning} or long short-term memory units (LSTM,
\cite{hochreiter1997long,gers2000learning}, can efficiently capture
long-term dependencies, the proposed network only needs a very small number of
convolutional layers.

We empirically validate the proposed model, to which we refer as a
convolution-recurrent network, on the eight large-scale document classification
tasks from \cite{zhang2015character}. We mainly compare the proposed model
against the convolutional network in \cite{zhang2015character} and show that it
is indeed possible to use a much smaller model to achieve the same level of
classification performance when a recurrent layer is put on top of the
convolutional layers.

\section{Basic Building Blocks: Neural Network Layers}

In this section, we describe four basic layers in a neural network that will be
used later to constitute a single network for classifying a document. 

\subsection{Embedding Layer}
\label{sec:emb_layer}

As mentioned earlier, each document is represented as a sequence of one-hot
vectors. A one-hot vector of the $i$-th symbol in a vocabulary is a binary
vector whose elements are all zeros except for the $i$-th element which is set
to one. Therefore, each document is a sequence of $T$ one-hot vectors $(\vx_1,
\vx_2, \ldots, \vx_T)$.

An {\em embedding layer} projects each of the one-hot vectors into a
$d$-dimensional continuous vector space $\RR^d$. This is done by simply
multiplying the one-hot vector from left with a weight matrix $\mW \in \RR^{d
\times |V|}$, where $|V|$ is the number of unique symbols in a vocabulary:
\begin{align*}
    \ve_t = \mW \vx_t.
\end{align*}
After the embedding layer, the input sequence of one-hot vectors becomes a
sequence of dense, real-valued vectors $(\ve_1, \ve_2, \ldots, \ve_T)$.

\subsection{Convolutional Layer}
\label{sec:conv_layer}

A {\em convolutional layer} consists of two stages. In the first stage, a set of
$d'$ filters of receptive field size $r$, $\mF \in \RR^{d' \times r}$, is
applied to the input sequence:
\begin{align*}
    \vf_t = \phi(\mF \left[ \ve_{t-(r/2)+1}; \ldots; \ve_{t}; \ldots,
    \ve_{t+(r/2)}\right]),
\end{align*}
where $\phi$ is a nonlinear activation function such as $\tanh$ or a rectifier.
This is done for every time step of the input sequence, resulting
 in a
sequence $F=(\vf_1, \vf_2, \ldots, \vf_T)$.

The resulting sequence $F$ is {\em max-pooled} with size $r'$:
\begin{align*}
    \vf'_t = \max\left( \vf_{(t-1)\times r'+1}, \ldots, \vf_{t \times r'}
    \right),
\end{align*}
where $\max$ applies for each element of the vectors, resulting in a sequence 
\[
    F' = (\vf'_1, \vf'_2, \ldots, \vf'_{T/r'}).
\]

\subsection{Recurrent Layer}
\label{sec:rec_layer}

A {\em recurrent layer} consists of a recursive function $f$ which takes as
input one input vector and the previous hidden state, and returns the new hidden
state:
\begin{align*}
    \vh_t = f(\vx_t, \vh_{t-1}),
\end{align*}
where $\vx_t \in \RR^{d}$ is one time step from the input sequence $(\vx_1, \vx_2, \ldots,
\vx_T)$. $\vh_0 \in \RR^{d'}$ is often initialized as an all-zero vector.

\paragraph{Recursive Function}

The most naive recursive function is implemented as
\begin{align*}
    \vh_t = \tanh\left( \mW_x \vx_t + \mU_h \vh_{t-1} \right),
\end{align*}
where $\mW_x \in \RR^{d' \times d}$ and $\mU_h \in \RR^{d' \times d'}$ are the
weight matrices. This naive recursive function however is known to suffer from
the problem of vanishing gradient
\cite{bengio1994learning,hochreiter2001gradient}. 

More recently it is common to use a more complicated function that learns to
control the flow of information so as to prevent the vanishing gradient and
allows the recurrent layer to more easily capture long-term dependencies. Long
short-term memory (LSTM) unit from \cite{hochreiter1997long,gers2000learning}
is a representative example. 

The LSTM unit consists of four sub-units--input, output, forget gates and
candidate memory cell, which are computed by
\begin{align*}
    &\vi_t = \sigma\left(\mW_i \vx_t + \mU_i \vh_{t-1}\right), \\
    &\vo_t = \sigma\left(\mW_o \vx_t + \mU_o \vh_{t-1}\right), \\
    &\vf_t = \sigma\left(\mW_f \vx_t + \mU_f \vh_{t-1}\right), \\
    &\tilde{\vc}_t = \tanh\left( \mW_c \vx_t + \mU_c \vh_{t-1} \right).
\end{align*}
Based on these, the LSTM unit first computes the memory cell:
\begin{align*}
    \vc_t = \vi_t \odot \tilde{\vc}_t + \vf_t \odot \vc_{t-1},
\end{align*}
and computes the output, or activation:
\begin{align*}
    \vh_t = \vo_t \odot \tanh(\vc_t).
\end{align*}

The resulting sequence from the recurrent layer is then 
\[
    (\vh_1, \vh_2, \ldots, \vh_T),
\]
where $T$ is the length of the input sequence to the layer.

\paragraph{Bidirectional Recurrent Layer}

One property of the recurrent layer is that there is imbalance in the amount of
information seen by the hidden states at different time steps. The earlier
hidden states only observe a few vectors from the lower layer, while the later
ones are computed based on the most of the lower-layer vectors. This can be
easily alleviated by having a bidirectional recurrent layer which is composed of
two recurrent layers working in opposite directions. This layer will return
two sequences of hidden states from the forward and reverse recurrent layers,
respectively.

\subsection{Classification Layer}

A {\em classification layer} is in essence a logistic regression classifier.
Given a fixed-dimensional input from the lower layer, the classification layer
affine-transforms it followed by a {\em softmax} activation function
\cite{bridle1990probabilistic} to compute the predictive probabilities for all
the categories. This is done by
\begin{align*}
    p(y=k | X) = \frac{\exp(\vw_k^\top \vx + b_k)}{\sum_{k'=1}^K \exp(\vw_{k'}^\top
    \vx + b_{k'})},
\end{align*}
where $\vw_k$'s and $b_k$'s are the weight and bias vectors. We assume there are
$K$ categories.

It is worth noting that this classification layer takes as input a {\em
fixed-dimensional} vector, while the recurrent layer or convolutional layer
returns a variable-length sequence of vectors (the length determined by the
input sequence). This can be addressed by either simply max-pooling
the vectors \cite{kim2014convolutional} over the time dimension (for both
convolutional and recurrent layers), taking the last hidden state (for recurrent
layers) or taking the last hidden states of the forward and reverse recurrent
networks (for bidirectional recurrent layers.)

\section{Character-Level Convolutional-Recurrent Network}

In this section, we propose a hybrid of convolutional and recurrent networks for character-level document classification. 

\subsection{Motivation}
\label{sec:motivation}
One basic motivation for using the convolutional layer is that it learns to
extract higher-level features that are invariant to local translation. By
stacking multiple convolutional layers, the network can extract higher-level,
abstract, (locally) translation-invariant features from the input sequence, in
this case the document, efficiently. 

Despite this advantage, we noticed that it requires many layers of convolution
to capture long-term dependencies, due to the locality of the convolution and
pooling (see Sec.~\ref{sec:conv_layer}.) This becomes more severe as the length
of the input sequence grows, and in the case of character-level modeling, it is
usual for a document to be a sequence of hundreds or thousands of characters.
Ultimately, this leads to the need for a very deep network having many
convolutional layers.

Contrary to the convolutional layer, the recurrent layer from
Sec.~\ref{sec:rec_layer} is able to capture long-term dependencies even when
there is only a single layer. This is especially true in the case of a
bidirectional recurrent layer, because each hidden state is computed based on
the whole input sequence. However, the recurrent layer is computationally more
expensive. The computational complexity grows linearly with respect to the
length of the input sequence, and most of the computations need to be done
sequentially. This is in contrast to the convolutional layer for which computations can be efficiently done in parallel.

Based on these observations, we propose to combine the convolutional and
recurrent layers into a single model so that this network can capture long-term dependencies in the document more efficiently for the task of classification.

\begin{figure}[t]
    \centering
    \begin{minipage}[b]{0.23\textwidth}
        \centering
        \includegraphics[width=0.98\columnwidth]{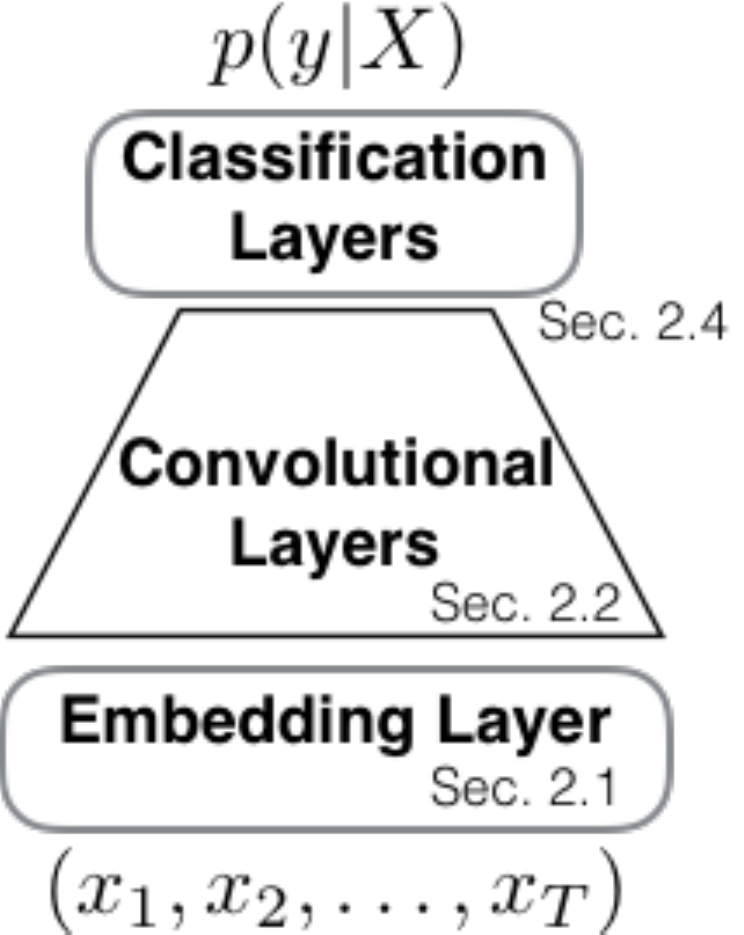}
        \\
        (a)
    \end{minipage}
    \hfill
    \begin{minipage}[b]{0.23\textwidth}
        \centering
        \includegraphics[width=0.98\columnwidth]{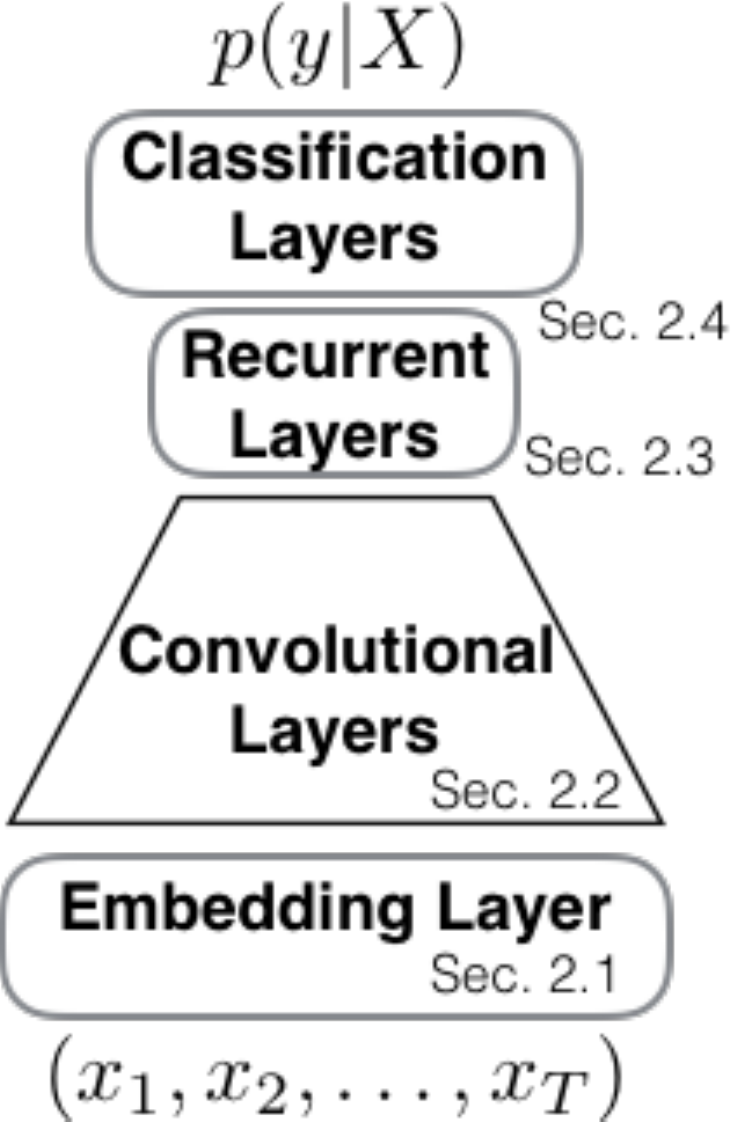}
        \\
        (b)
    \end{minipage}
    \caption{Graphical illustration of (a) the convolutional network and (b) the proposed convolution-recurrent
    network for character-level document classification.}
    \label{fig:models}
\end{figure}

\subsection{Model Description}

The proposed model, to which we refer as a convolution-recurrent network
(ConvRec), starts with a one-hot sequence input
\[
    X = (\vx_1, \vx_2, \ldots, \vx_T).
\]
This input sequence is turned into
a sequence of dense, real-valued vectors 
\[
    E = (\ve_1, \ve_2, \ldots, \ve_T)
\]
using the {\em embedding layer} from
Sec.~\ref{sec:emb_layer}. 

We apply multiple {\em convolutional layers} (Sec.~\ref{sec:conv_layer}) to $E$ to get a
shorter sequence of feature vectors:
\[
    F = (\vf_1, \vf_2, \ldots, \vf_{T'}).
\]
This feature vector is then fed into a {\em bidirectional recurrent layer}
(Sec.~\ref{sec:rec_layer}), resulting in two sequences
\begin{align*}
    &H_{\text{forward}} = (\ora{\vh}_1, \ora{\vh}_2, \ldots, \ora{\vh}_{T'}), \\
    &H_{\text{reverse}} = (\ola{\vh}_1, \ola{\vh}_2, \ldots, \ola{\vh}_{T'}). 
\end{align*}
We take the last hidden states of both directions and concatenate
them to form a fixed-dimensional vector:
\[
    \vh = \left[ \ora{\vh}_{T'}; \ola{\vh}_{1}\right].
\]

Finally, the fixed-dimensional vector $\vh$ is fed into the {\em classification layer} to compute the predictive probabilities $p(y=k | X)$ of all the
categories $k=1,\ldots,K$ given the input sequence $X$. 

See Fig.~\ref{fig:models}~(b) for the graphical illustration of the proposed
model.

\begin{table*}
\centering
\begin{tabular}{|l|cccc|}
\hline \bf Data set & \bf Classes & \bf Task & \bf Training size & \bf Test size \\ 
    \hline
    \hline
AG's news & 4 &news categorization & 120,000 & 7,600 \\
Sogou news & 5 &news categorization & 450,000 & 60,000\\
DBPedia & 14 &ontology classification & 560,000 & 70,000\\
Yelp review polarity & 2 &sentiment analysis & 560,000 & 38,000\\
Yelp review full & 5 &sentiment analysis & 650,000 & 50,000\\
Yahoo! Answers & 10 &question type classification & 1,400,000 & 60,000\\
Amazon review polarity & 2 &sentiment analysis & 3,600,000 & 400,000\\
Amazon review full & 5 &sentiment analysis & 3,000,000 & 650,000\\
\hline
\end{tabular}
\vskip 0.1in
\caption{\label{data-table} Data sets summary.}
\end{table*}

\subsection{Related Work}

\paragraph{Convolutional network for document classification}

The convolutional networks for document classification, proposed earlier in
\cite{kim2014convolutional,zhang2015character} and illustrated in
Fig.~\ref{fig:models}~(a), is almost identical to the
proposed model. One major difference is the lack of the recurrent layer in their
models. Their model consists of the embedding layer, a number of convolutional
layers followed by the classification layer only.

\paragraph{Recurrent network for document classification}

Carrier and Cho in \cite{Carrier2014} give a tutorial on using a recurrent
neural network for sentiment analysis which is one type of document
classification. Unlike the convolution-recurrent network proposed in this paper,
they do not use any convolutional layer in their model. Their model starts with
the embedding layer followed by the recurrent layer. The hidden states from the
recurrent layer are then averaged and fed into the classification layer. 

\paragraph{Hybrid model: Conv-GRNN}

Perhaps the most related work is the convolution-gated recurrent neural net
(Conv-GRNN) from \cite{tang2015document}. They proposed a hierarchical
processing of a document. In their model, either a convolutional network or a
recurrent network is used to extract a feature vector from each sentence, and
another (bidirectional) recurrent network is used to extract a feature vector of
the document by reading the sequence of sentence vectors. This document vector
is used by the classification layer.

The major difference between their approach and the proposed ConvRec is in the
purpose of combining the convolutional network and recurrent network.  In their
model, the convolutional network is strictly constrained to model each sentence,
and the recurrent network to model inter-sentence structures. On the other hand,
the proposed ConvRec network uses a recurrent layer in order to assist the
convolutional layers to capture long-term dependencies (across the whole
document) more efficiently. These are orthogonal to each other, and it is
possible to plug in the proposed ConvRec as a sentence feature extraction module
in the Conv-GRNN from \cite{tang2015document}. Similarly, it is possible to use
the proposed ConvRec as a composition function for the sequence of sentence
vectors to make computation more efficient, especially when the input document
consists of many sentences.

\paragraph{Recursive Neural Networks}

A recursive neural network has been applied to sentence classification earlier
(see, e.g., \cite{socher2013recursive}.) In this approach, a composition
function is defined and recursively applied at each node of the parse tree of an
input sentence to eventually extract a feature vector of the sentence.  This
model family is heavily dependent on an external parser, unlike all the other
models such as the ConvRec proposed here as well as other related models
described above. It is also not trivial to apply the recursive neural network to
documents which consist of multiple sentences. We do not consider this family of
recursive neural networks directly related to the proposed model.

\begin{table*}
\centering
\begin{tabular}{|l|ccc|cccc|c|}
    \hline
& \multicolumn{3}{c|}{\textbf{Embedding Layer}} & \multicolumn{4}{c|}{\textbf{Convolutional Layer}} & \textbf{Recurrent Layer} \\ 
\bf Model & \multicolumn{3}{c|}{Sec.~\ref{sec:emb_layer}} & \multicolumn{4}{c|}{Sec.~\ref{sec:conv_layer}} & Sec.~\ref{sec:rec_layer} \\ 
&\hspace{2ex} & $|V|$ & $d$ & $d'$ & $r$ & $r'$ & $\phi$ & $d'$\\ 
\hline
\hline
C2R1D$D$ && \multirow{4}{*}{96} & \multirow{4}{*}{8} &\multirow{4}{*}{$D$} &5,3 &2,2 &\multirow{4}{*}{ReLU} & \multirow{4}{*}{$D$}\\
C3R1D$D$ && & & &5,5,3 &2,2,2 & & \\
C4R1D$D$ && & & &5,5,3,3 &2,2,2,2 & & \\
C5R1D$D$ && & & &5,5,3,3,3 &2,2,2,1,2 & & \\
\hline
\end{tabular}
\vskip 0.1in
\caption{\label{model-table} Different architectures tested in this paper.}
\end{table*}

\section{Experiment Settings}

\subsection{Task Description}

We validate the proposed model on eight large-scale document classification tasks from \cite{zhang2015character}. The sizes of the data sets range from 200,000 to 4,000,000 documents. These tasks include sentiment analysis (Yelp reviews, Amazon reviews), ontology classification (DBPedia), question type classification (Yahoo! Answers), and news categorization (AG's news, Sogou news). 
\paragraph{Data Sets}
A summary of the statistics for each data set is listed in Table \ref{data-table}. There are equal number of examples in each class for both training and test sets. DBPedia data set, for example, has 40,000 training and 5,000 test examples per class. For more detailed information on the data set construction process, see \cite{zhang2015character}.

\subsection{Model Settings}
Referring to Sec. \ref{sec:emb_layer}, the vocabulary $V$ for our experiments consists of 96 characters including all upper-case and lower-case letters, digits, common punctuation marks, and spaces. Character embedding size $d$ is set to 8. 

As described in Sec.~\ref{sec:motivation}, we believe by adding recurrent layers, one can effectively reduce the number of convolutional layers needed in order to capture long-term dependencies. Thus for each data set, we consider models with two to five convolutional layers. Following notations in Sec. \ref{sec:conv_layer}, each layer has $d'=128$ filters. For AG's news and Yahoo! Answers, we also experiment larger models with 1,024 filters in the convolutional layers. Receptive field size $r$ is either five or three depending on the depth. Max pooling size $r'$ is set to 2. Rectified linear units (ReLUs, \cite{glorot2011relu}) are used as activation functions in the convolutional layers. The recurrent layer (Sec. \ref{sec:rec_layer}) is fixed to a single layer of bidirectional LSTM for all models. Hidden states dimension $d'$ is set to 128. More detailed setups are described in Table \ref{model-table}.

Dropout \cite{srivastava2014dropout} is an effective way to regularize deep neural networks. We apply dropout after the last convolutional layer as well as after the recurrent layer. Without dropout, the inputs to the recurrent layer $\vx_t$'s are
\[\vx_t=\vf'_t\]
where $\vf'_t$ is the $t$-th output from the last convolutional layer defined in Sec. \ref{sec:conv_layer}. After adding dropout, we have
\[r^i_t\sim\operatorname{Bernoulli}(p)\]
\[\vx_t=\vr_t\odot\vf'_t\]
$p$ is the dropout probability which we set to 0.5; $r^i_t$ is the $i$-th component of the binary vector $\vr_t\in\RR^{d'}$.

\begin{table*}
    \centering
    \resizebox{\textwidth}{!}{%
    \begin{tabular}{l || c c | c c c | c c c }
        & & &
        \multicolumn{3}{c|}{Our Model} &
        \multicolumn{3}{c}{\cite{zhang2015character}} \\
        Data set & \# Ex. & \# Cl. & 
        Network & \# Params & Error (\%) &
        Network & \# Params & Error (\%) \\
        \hline \hline
        AG        & 120k & 4  & C2R1D1024 & 20M & 8.39/{\bf 8.64} & C6F2D1024 & 27M & -/9.85 \\
        Sogou     & 450k & 5  & C3R1D128  & .4M & 4.82/{\bf 4.83} & C6F2D1024$^\star$ & 27M & -/4.88 \\
        DBPedia   & 560k & 14 & C2R1D128  & .3M & 1.46/{\bf 1.43} & C6F2D1024 & 27M & -/1.66 \\
        Yelp P.   & 560k & 2  & C2R1D128  & .3M & 5.50/5.51 & C6F2D1024 & 27M & -/{\bf 5.25} \\
        Yelp F.   & 650k & 5  & C2R1D128  & .3M & 38.00/{\bf 38.18} & C6F2D1024 & 27M & -/38.40 \\
        Yahoo A.  & 1.4M & 10 & C2R1D1024 & 20M & 28.62/{\bf 28.26} & C6F2D1024$^\star$ & 27M& -/29.55 \\
        Amazon P. & 3.6M & 2  & C3R1D128  & .4M & 5.64/5.87 & C6F2D256$^\star$  & 2.7M& -/{\bf 5.50} \\
        Amazon F. & 3.0M & 5  & C3R1D128  & .4M & 40.30/40.77 & C6F2D256$^\star$ & 2.7M& -/{\bf 40.53} \\
    \end{tabular}}
\vskip 0.1in
\caption{
    Results on character-level document classification. C$C$R$R$F$F$D$D$ refers to a
    network with $C$ convolutional layers, $R$ recurrent layers, $F$
    fully-connected layers and $D$
    dimensional feature vectors. $\star$ denotes a model which does not distinguish
    between lower-case and upper-case letters. We only considered the
    character-level models without using Thesaraus-based data augmentation.  We report both the validation and test errors.
    In our case, the network architecture for each dataset was selected based on
the validation errors. The numbers of parameters are approximate.} 
\label{result-table} 
\end{table*}

\subsection{Training and Validation}

For each of the data sets, we randomly split the full training examples into training and validation. The validation size is the same as the corresponding test size and is balanced in each class.

The models are trained by minimizing the following regularized negative log-likelihood or cross entropy loss. $X$'s and $y$'s are document character sequences and their corresponding observed class assignments in the training set $D$. $\vw$ is the collection of model weights. Weight decay is applied with $\lambda=5\times10^{-4}$.
\[l=-\sum_{X,y\in D}\log(p(y|X))+\frac{\lambda}{2}\|\vw\|^2\]

We train our models using AdaDelta \cite{zeiler2012adadelta} with $\rho=0.95$, $\epsilon=10^{-5}$ and a batch size of 128. Examples are padded to the longest sequence in each batch and masks are generated to help identify the padded region. The corresponding masks of the outputs from convolutional layers can be computed analytically and are used by the recurrent layer to properly ignore padded inputs. The gradient of the cost function is computed with backpropagation through time (BPTT, \cite{werbos1990bptt}). If the gradient has an L2 norm larger than 5, we rescale the gradient by a factor of $\frac{5}{\|\vg\|_2}$. i.e.
\[\vg_c=\vg\cdot \min\left(1, \frac{5}{\|\vg\|_2}\right)\]
where $\vg=\frac{\text{d}l}{\text{d}\vw}$ and $\vg_c$ is the clipped gradient.

Early stopping strategy is employed to prevent overfitting. Before training, we set an initial $patience$ value. At each epoch, we calculate and record the validation loss. If it is lower than the current lowest validation loss by 0.5\%, we extend $patience$ by two. Training stops when the number of epochs is larger than $patience$. We report the test error rate evaluated using the model with the lowest validation error.

\section{Results and Analysis}
Experimental results are listed in Table \ref{result-table}. We compare to the best character-level convolutional model without data augmentation from \cite{zhang2015character} on each data set. Our model achieves comparable performances for all the eight data sets with significantly less parameters. Specifically, it performs better on AG's news, Sogou news, DBPedia, Yelp review full, and Yahoo! Answers data sets. 

\begin{figure*}[t]
    \centering
    \begin{minipage}{0.49\textwidth}
    \centering
    \includegraphics[width=0.98\columnwidth]{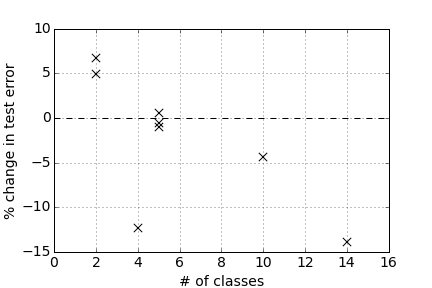}
(a)
\end{minipage}
\hfill
\begin{minipage}{0.49\textwidth}
\centering
    \includegraphics[width=0.98\columnwidth]{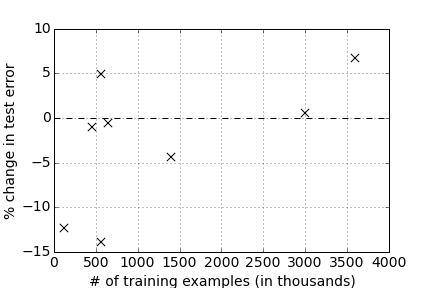}
(b)
\end{minipage}
    \caption{Relative test performance of the proposed model compared to the convolution-only model w.r.t. (a) the number of classes and (b) the size of training set. Lower is better.}
    \label{fig:analysis}
\end{figure*}

\paragraph{Number of classes} Fig. \ref{fig:analysis} (a) shows how relative performance of our model changes with respect to the number of classes. It is worth noting that as the number of classes increases, our model achieves better results compared to convolution-only models. For example, our model has a much lower test error on DBPedia which has 14 classes, but it scores worse on Yelp review polarity and Amazon review polarity both of which have only two classes.  Our conjecture is that more detailed and complete information needs to be preserved from the input text for the model to assign one of many classes to it. The convolution-only model likely loses detailed local features because it has more pooling layers. On the other hand, the proposed model with less pooling layers can better maintain the detailed information and hence performs better when such needs exist.

\paragraph{Number of training examples} Although it is less significant, Fig. \ref{fig:analysis} (b) shows that the proposed model generally works better compared to the convolution-only model when the data size is small. Considering the difference in the number of parameters, we suspect that because the proposed model is more compact, it is less prone to overfitting. Therefore it generalizes better when the training size is limited.

\paragraph{Number of convolutional layers}
An interesting observation from our experiments is that the model accuracy does not always increase with the number of convolutional layers. Performances peak at two or three convolutional layers and decrease if we add more to the model. As more convolutional layers produce longer character n-grams, this indicates that there is an optimal level of local features to be fed into the recurrent layer. Also, as discussed above, more pooling layers likely lead to the lost of detailed information which in turn affects the ability of the recurrent layer to capture long-term dependencies.

\paragraph{Number of filters}
We experiment large models with 1,024 filters on AG's news and Yahoo! Answers data sets. Although adding more filters in the convolutional layers does help with the model performances on these two data sets, the gains are limited compared to the increased number of parameters. Validation error improves from 8.75\% to 8.39\% for AG's news and from 29.48\% to 28.62\% for Yahoo! Answers at the cost of a 70 times increase in the number of model parameters. 

Note that in our model we set the number of filters in the convolutional layers to be the same as the dimension of the hidden states in the recurrent layer. It is possible to use more filters in the convolutional layers while keeping the recurrent layer dimension the same to potentially get better performances with less sacrifice of the number of parameters.

\section{Conclusion}

In this paper, we proposed a hybrid model that processes an input sequence of characters with a number of convolutional layers followed by a single recurrent layer. The proposed model is able to encode documents from character level capturing sub-word information.

We validated the proposed model on eight large scale document classification tasks. The model achieved comparable results with much less convolutional layers compared to the convolution-only architecture. We further discussed several aspects that affect the model performance. The proposed model generally performs better when number of classes is large, training size is small, and when the number of convolutional layers is set to two or three.

The proposed model is a general encoding architecture that is not limited to document classification tasks or natural language inputs. For example, \cite{ChenBPMY15,VisinKCBMC15} combined convolution and recurrent layers to tackle image segmentation tasks; \cite{sainath2015speech} applied a similar model to do speech recognition. It will be interesting to see future research on applying the architecture to other applications such as machine translation and music information retrieval. Using recurrent layers as substitutes for pooling layers to potentially reduce the lost of detailed local information is also a direction that worth exploring.

\section*{Acknowledgments}
This work is done as a part of the course DS-GA 1010-001 Independent Study in Data Science at the Center for Data Science, New York University.

\bibliography{naaclhlt2016}
\bibliographystyle{naaclhlt2016}

\end{document}